\documentclass{article}

\usepackage{arxiv}

\usepackage[utf8]{inputenc} 
\usepackage[T1]{fontenc}    
\usepackage{hyperref}       
\usepackage{url}            
\usepackage{booktabs}       
\usepackage{amsfonts}       
\usepackage{nicefrac}       
\usepackage{microtype}      
\usepackage{lipsum}
\usepackage{graphicx}
\usepackage{graphics}
\graphicspath{ {./images/} }

\usepackage{amsmath} 
\usepackage{amssymb}  
\usepackage{booktabs}
\usepackage{siunitx}
\usepackage{multirow}
\usepackage[dvipsnames]{xcolor}
\usepackage{hyperref}
\usepackage{tabularx}

\usepackage[export]{adjustbox}
\usepackage{algorithm}
\usepackage{algpseudocode}
\usepackage{marvosym}

\title{Semi-Supervised Novelty Detection for Precise Ultra-Wideband Error Signal Prediction}

\author{Umberto~Albertin$^{1}$, Alessandro~Navone$^{1}$, Mauro~Martini$^{1}$ and Marcello Chiaberge$^{1}$ 
\thanks{$^{1}$ Department of Electronics and Telecommunications, Politecnico di Torino, 10129, Torino, Italy. \tt\footnotesize \{firstname.lastname\}@polito.it}}

\author{
Umberto Albertin\\
  Department of Electronics and Telecommunications \\
  Politecnico di Torino\\
  Torino, TO, 10129 \\
  \texttt{umberto.albertin@polito.it} \\
  \And
 Alessandro Navone \\
  Department of Electronics and Telecommunications \\
  Politecnico di Torino\\
  Torino, TO, 10129 \\
  \texttt{alessandro.navone@polito.it} \\
   \And
 Mauro Martini \\
  Department of Electronics and Telecommunications \\
  Politecnico di Torino\\
  Torino, TO, 10129 \\
  \texttt{mauro.martini@polito.it} \\
   \And
 Marcello Chiaberge \\
  Department of Electronics and Telecommunications \\
  Politecnico di Torino\\
  Torino, TO, 10129 \\
  \texttt{marcello.chiaberge@polito.it} \\
}

\begin{document}
\maketitle
\begin{abstract}
Ultra-Wideband (UWB) technology is an emerging low-cost solution for localization in a generic environment. However, UWB signal can be affected by signal reflections and non-line-of-sight (NLoS) conditions between anchors; hence, in a broader sense, the specific geometry of the environment and the disposition of obstructing elements in the map may drastically hinder the reliability of UWB for precise robot localization. 
This work aims to mitigate this problem by learning a map-specific characterization of the UWB quality signal with a fingerprint semi-supervised novelty detection methodology. An unsupervised autoencoder neural network is trained on nominal UWB map conditions, and then it is used to predict errors derived from the introduction of perturbing novelties in the environment. This work poses a step change in the understanding of UWB localization and its reliability in evolving environmental conditions. The resulting performance of the proposed method is proved by fine-grained experiments obtained with a visual tracking ground truth.
\end{abstract}

\section{Introduction}\label{sec:intro}

Service mobile robots are rapidly emerging as the new frontier of automation in our daily life activities, from domestic and health assistance \cite{ eirale2022human, tamantini2021robotic} to precision agriculture \cite{navone2023rows, martini2022} and inspection \cite{gehring2021anymal}. Localization technologies are the primary asset to enhance the future development and reliability of autonomous mobile robots \cite{alatise2020review}. Wheel odometry and visual odometry are standard methods that have been widely investigated in the last decade, although struggling when dealing with long paths and poor repetitive features \cite{navone2023online, wang2020approaches, gupta2020corridor}.

\begin{figure}
    \centering
    \includegraphics[width=0.6\linewidth]{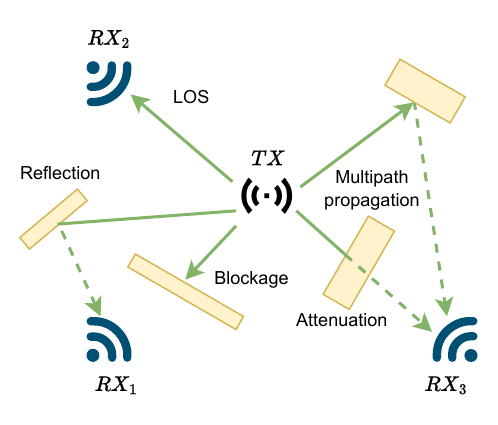}
    \caption{UWB signal attenuation types depending on the obstacles within and around the environment. A novel method for UWB Error Prediction with Semi-Supervised Novelty Detection (EPSNoDe) is proposed to identify the map areas where the dynamic nature of the environment may affect the UWB signal.}
    \label{fig:schema-intro}
\end{figure}

Among localization technologies, Ultra-Wideband (UWB) has recently emerged as a promising low-cost candidate to localize mobile robots and devices in GPS-denied environments \cite{elsanhoury2022precision}.
UWB technology operates by transmitting extremely short-duration pulses, allowing for precise time-of-arrival estimations and, thus, highly accurate distance measurements. However, despite its potential, UWB is still affected by primary sources of uncertainty as the Non-Line of Sight (NLoS) propagation condition and multipath reflections, as shown in Fig \ref{fig:schema-intro}, which can significantly degrade the localization accuracy. These phenomena occur when the signal cannot travel directly from the transmitter to the receiver or when it bounces off surfaces, respectively. This leads to a biased distance estimation between devices, hindering the possibility of performing autonomous navigation tasks. Moreover, the continuous changes in the environment where the robot operates increase the difficulty of identifying areas where the UWB localization system's reliability can drop.

\subsection{Related works}
Several studies have attempted to address these issues. For instance, machine learning and deep learning techniques have been proposed to identify and mitigate NLoS conditions, framing the problem as both a classification and a regression task \cite{che2022feature, angarano2021robust}. Others have explored using additional information, such as inertial measurements \cite{goudar2021online} or map knowledge \cite{wang2021high}, to improve UWB-based localization. In some cases, the position of a receiver was directly estimated from raw data without the necessity to identify a-priori NLoS conditions. The aforementioned techniques have demonstrated a high accuracy even in complex environments, where it is not always possible to guarantee a Line of Sight (LoS) condition \cite{nitsoo2019deep}.  Despite these efforts, a comprehensive solution that can effectively handle real-world environments' diverse and dynamic nature is still lacking.

In this work, we focus on the central role of the environment on the overall performance of the UWB, considering the dynamic nature of UWB error strictly coupled with the introduction of anomalies in the map. To this aim, novelty detection is a set of machine learning techniques capable of identifying new or unknown patterns in data distribution starting from nominal learned data \cite{gruhl2021novelty}. In the context of UWB signals, novelty detection can identify NLoS and multipath conditions as novel situations, as well as significative changes in the environment, providing the necessary position-specific uncertainty information for a reliable localization. 
Autoencoders are frequently employed in novelty detection applications due to their ability to acquire a condensed representation of data \cite{del2022novelty}\cite{zhao2023novelty}. Within the Ultra-Wideband (UWB) technology, autoencoders find applications in several domains, such as signal reliability estimation and correction. Variational autoencoders have effectively assigned anomaly scores to the channels of Time-of-Flight (ToF)-based systems. This showcases notable generalization capabilities, permitting the incorporation of the anomaly scores into an Extended Kalman Filter-based position tracking system \cite{stahlke2022estimating}. Furthermore, autoencoders have been leveraged to enhance the precision of ranging measurements, particularly in the context of localization \cite{fontaine2020edge}.

\subsection{Contributions}
In this research, we proposed a semi-supervised novelty detection methodology to precisely characterize UWB error within a specific environment. Our approach integrates the novelty detection framework with deep learning, opting for unsupervised autoencoders as the underlying neural network architecture. This investigation focuses on discerning how subtle environmental alterations can contribute to a degradation of UWB signal quality.

Conversely, delineating nominal environmental conditions is considered an indirect introduction of human supervision into the data handling process. Therefore, we classify our proposed methodology as semi-supervised. A comprehensive experimental campaign has been executed to meticulously gather fine-grained measurements of UWB signals alongside ground truth position data for testing purposes, facilitated by a Vicon motion capture system. By leveraging the advantages of novelty identification with autoencoders, this work sheds new light on the reliability of UWB-based localization, considering a general characterization of environmental changes. 

\section{Methodology}\label{sec:methodology}

For the development of this work, we considered an environment with fixed UWB anchors and a moving tag. A dataset is then collected to gather a database wherein  UWB-specific signal characteristics are systematically recorded at predetermined reference points in the environment. This represents the so-called offline phase. The development of the localization algorithm is framed as a mathematical function. This consists of minimizing the localization error through an optimization process to identify a reference point given a set of signal characteristics during the online phase. This localization algorithm, employed in wireless technologies, is commonly called the fingerprint technique. This technique, valid in static environments, is strongly influenced by changes in the environment, which affect the characteristics of the signal. The novel method proposed in this work aims to identify when the signal is no longer reliable by employing deep learning-based novelty identification techniques. We refer to the methodology proposed as UWB Error Prediction with Semi-Supervised Novelty Detection (EPSNoDe).

\subsection{Novelty Identification}

In our paper, novelty detection has a crucial role in warning zone identification, recognizing areas where the localization signal appears to be degraded due to environmental conditions. The novelty detection framework is characterized by a training process exclusively involving nominal data. Since the model must recognize changes in the data during the inference phase, these alterations should not be included in the training phase. This approach intentionally induces overfitting in the model, rendering it specific to the training data while compromising its generalization capabilities. 

Since the fingerprint approach is employed, the novelty framework is consistently trained with similar features for each position in the grid map. The core of this study focuses on identifying environmental changes using UWB signals. The model assimilates the signal pattern as a function of the robot's position in the grid map, identifying if and where a new pattern emerges. Introducing an obstacle in the environment induces UWB reflections and NLoS conditions, causing a distorted signal compared to the original. Consequently, the model detects this distortion, indicating a novelty in the specific position where the altered signal is detected.

The novelty identification models use a reconstruction error to understand if the signal's features are new with respect to the training ones. In this paper, two errors are computed for detecting the novelty: the first one is related to a single UWB anchor by computing a difference between the real features and the reconstructed ones, as reported in the following equation:

\begin{equation} \label{eq:err}
   e = |\hat{y} - y|
\end{equation} 

where $e$ represents the error, $\hat{y}$ represents the predicted sample, and $y$ represents the real one. The second error is computed by using the errors of $i_{th}$ anchors and summing them as a multidimensional distance, as reported in the following equation:

\begin{equation} \label{eq:toterr}
   e_{total} = \sqrt{e_1^2 + e_2^2 + ... + e_n^2}
\end{equation} 

where $e_{total}$ is the total error computed for a single prediction, $e_i$ is the error of the $i_{th}$ anchor prediction computed by using Eq. \ref{eq:err}, and $n$ is the total number of anchors used for the robot's localization. 

\subsection{Neural Network Architecture}\label{subsec:nn}
The architecture considered for the purpose outlined in this paper is an autoencoder. Autoencoders are powerful tools for semi-unsupervised learning, showing the ability to learn compact and meaningful representations of input data \cite{pinaya2020autoencoders}. Their structure comprises two main components: the encoder and the decoder. The encoder transforms high-dimensional input data into a latent space, usually characterized by a lower dimensionality, capturing essential features. Subsequently, the decoder reconstructs the original input from this condensed representation.
\begin{figure}[h]
    \centering
    \includegraphics[width=0.80\linewidth]{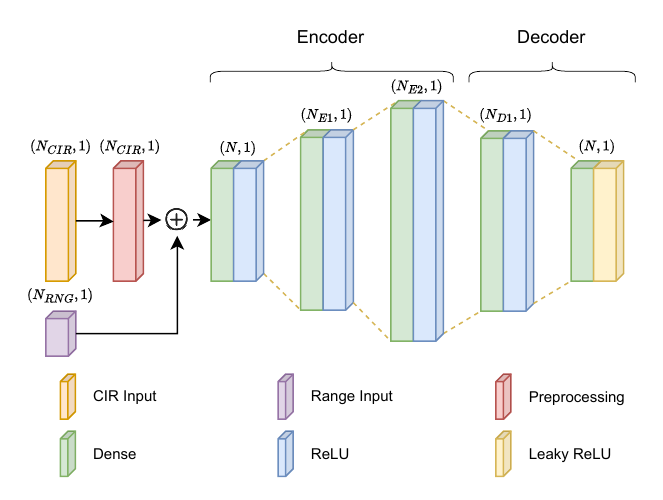}
    \caption{Architecture of the proposed UWB EPSNoDe model - It consists of an overcomplete autoencoder ($N < N_{E1} < N_{E2}$ and $N_{D1} > N$). The legend illustrates all the layer types. The input dimension varies according to the input type.}
    \label{fig:autoencoder}
\end{figure}
Formally, let $X \in \mathbb{R} ^ n$ represent the input data, where $n$ denotes the dimensionality of each sample. The autoencoder comprises the encoder, $f_{enc}: \mathbb{R}^{n} \rightarrow \mathbb{R}^{p}$, and the decoder $f_{dec}: \mathbb{R}^{p} \rightarrow \mathbb{R}^{n}$. This facilitates the transformation of input data into a latent space $h = f_{enc}(x)$ and a subsequent reconstruction $X_{recon} = f_{dec}(h)$. Extending the autoencoder paradigm, \textit{overcomplete} autoencoders deliberately operate with a latent space of higher dimensionality ($p > n$). This intentional over-parametrization in overcomplete autoencoders allows capturing a richer set of features \cite{bengio2009learning}.

In this work, an overcomplete autoencoder architecture is employed, as shown in Fig. \ref{fig:autoencoder}. The encoder stage consists in three dense layers with dimensions $N$, $N_{E1}$ and $N_{E2}$, where $N = n$, $N_{E2} = p$ and $N < N_{E1} < N_{E2}$. The three layers' activation functions consist of \textit{ReLU} functions. Subsequently, the decoder consists of two layers with dimensions $N_{D1}$ and $N$, where $N_{D1} > N$, restoring the initial dimension. The first decoder layer utilizes a ReLU activation function. In contrast, the second layer employs a Leaky ReLU activation, introducing a slight negative slope to handle potential dead neurons and enhance the robustness of the model. The overall architecture is trained to employ a Mean Squared Error (MSE) loss function. Nonetheless, different UWB data selection and pre-processing strategies are explored to extract relevant information from the sensor raw data: \textit{EPSNoDe\textsubscript{RNG}} configuration considers only the anchors' ranges as input for the network; \textit{EPSNoDe\textsubscript{MA}} incorporates both the anchors' ranges and the first 6 peaks of the Moving Average applied to the Channel Input Response (CIR) signal of the UWB; lastly, \textit{EPSNoDe\textsubscript{PCA}} incorporates the anchors' ranges and a CIR signal reduced by applying a Principal Component Analysis (PCA) to the signal itself.

\subsection{Position-specific Error Prediction}

The final stage of the framework involves identifying unreliable zones of the grid map for UWB localization through the utilization of output data predicted by the \textit{EPSNoDe} model. The reconstruction errors computed are used as key indicators to identify if a specific grid point is anomalous with respect to the initial environment. Detecting an anomaly in a grid zone implies the possibility of an environmental change occurring in proximity to the point itself. A possible example of the usage of the error could be to impose a threshold, giving a warning when this threshold is exceeded. In that position the autonomous robot recognizes that the localization algorithm is influenced, prompting the potential implementation of strategies in subsequent steps to enhance localization precision using UWB. These strategies (e.g., error mitigation or compensation) and the selection of a possible error threshold are out of the scope of this paper. 

\section{Tests and Results}\label{sec:results}
Several experiments are conducted to evaluate the proposed framework's performance and functionality. 

\begin{figure}[t]
    \centerline{\includegraphics[width=0.7\columnwidth]{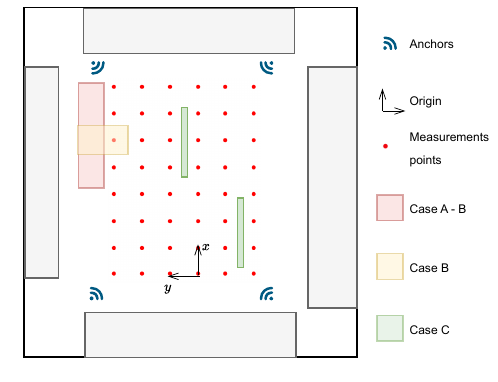}}
    \caption{Sketch of the environment used to test the algorithm. Three scenarios of novelty can be identified: A, B, C.}
    \label{fig:path}
\end{figure}

\subsection{Experimental Setup}
The environment used for the framework performance evaluation is an office room at the PIC4SeR center. This room is characterized by several desks, chairs, and wardrobes placed around it. A sub-environment is obtained in a free rectangle in this room, where the robot can move. The planned path is a rectangular grid with a 50 cm mesh size along $x$ and $y$ directions. Three different experimental scenarios are considered, as shown in Fig. \ref{fig:path}. In Case A, a metal plate is positioned beside the top-right corner of the grid map. In Case B, the same scenario with a wooden bridge and a metal plate placed in the top-right corner creates NLoS conditions at several points of the grid map. The last condition, Case C, shows two obstacles placed within the grid map. In all the experiments, the rest of the environment has been left unchanged. Four distinct datasets, encompassing clear and perturbed scenarios, are gathered throughout the experiment phase. The obstacles added to each environment are higher than the robot, thereby establishing NLoS conditions by obstructing anchors in certain grid sections during the measurement phase.

\subsection{Experiments and Results}
\begin{figure}[t]
    \centerline{\includegraphics[width=\linewidth]{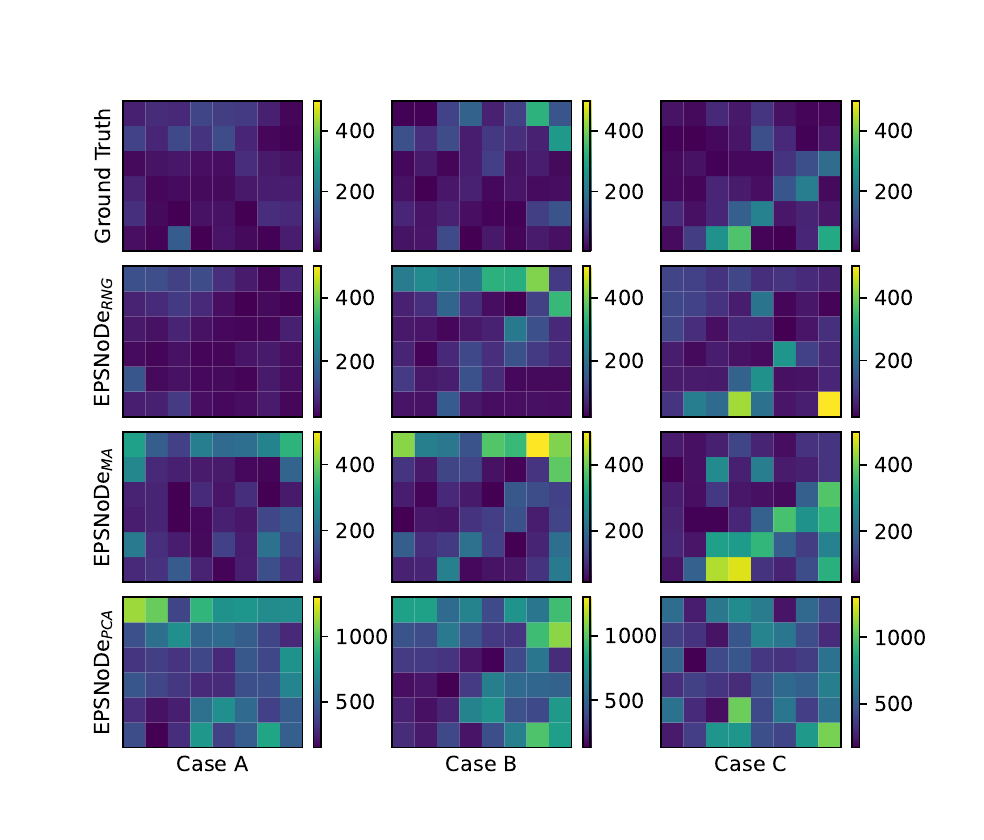}}
    \caption{The total error is computed for each test conducted. Each row corresponds to the architecture type applied to the dataset, and each column corresponds to the dataset type employed in the test. \textit{EPSNoDe\textsubscript{RNG}} uses only ranging distances, \textit{EPSNoDe\textsubscript{MA}} involves the Moving Average of the CIR along with the ranging distances, and \textit{EPSNoDe\textsubscript{PCA}} involves the application of Principal Component Analysis to the CIR along with the ranging distances.}
    \label{fig:errors}
\end{figure}

An extensive experimental session is carried out to demonstrate the effectiveness of the method in characterizing the quality of the UWB signal in the environment. Moreover, we investigate the role of UWB data selection in the learning process. More precisely, we defined three different pre-processing operations for the proposed autoencoder architecture described in \ref{subsec:nn}. Each network configuration is tested on scenarios A, B, and C, and the results obtained are shown in Fig. \ref{fig:errors}. Each quadrant shows the total error computed for the related position, where a bright quadrant indicates a high error. 

The first model analyzed is the \textit{EPSNoDe\textsubscript{RNG}} which specifically detects novelty based only on ranging distances. In this scenario, the model employs the minimum number of features compared to other models. Using four anchors, the features are limited to four, corresponding to the ranging values of each anchor. As it can be noticed, the model detects the presence of the obstacles in both Case B and C where the obstacles are positioned within the map. The obstacle in Case A is less visible using only ranging distance, possibly owing to the persistent LoS conditions throughout the measurement phase. By exclusively considering ranging distance, potential reflections of the CIR are neglected in the analysis. On the opposite side, the \textit{EPSNoDe\textsubscript{PCA}} model exhibits a contrasting performance: in this instance, a more extensive set of features is considered compared to other studies. The original CIR comprises 152 samples multiplied by the number of anchors, resulting in 608 features for our case with 4 anchors. The PCA model used reduces the feature dimension, retaining 90\% of the variance of the initial data, reducing the features to only 68 elements. Adding the four UWB ranges to the array results in 72 features as input for the framework. Due to this elevated number of features, this architecture encounters some difficulties in learning the patterns for each point of the grid map. The results obtained exhibit a considerable level of uncertainty in the overall map. 

The most promising outcomes in this study are associated with the \textit{EPSNoDe\textsubscript{RNG}} and \textit{EPSNoDe\textsubscript{MA}} where a two-period moving average is applied to each anchor's CIR to filter the raw signal and its reflections. In the \textit{EPSNoDe\textsubscript{MA}} approach, the framework considers the first six peaks of the CIR, along with the ranging distances, as inputs. 

The obstacles in cases B and C are clearly visible on the map. Case A related to the obstacle beside the map, is harder to recognize. Only the \textit{EPSNoDe\textsubscript{MA}} detects some novelties in the top row of the map. These distinctive elements are solely recognized due to the reflections of the UWB signal interacting with the obstacle beside the map. Incorporating the first six peaks of the CIR into the range distances, significant signal reflections are integrated and detected as novelties.

\begin{figure}[t]
    \centerline{\includegraphics[width=0.9\linewidth]{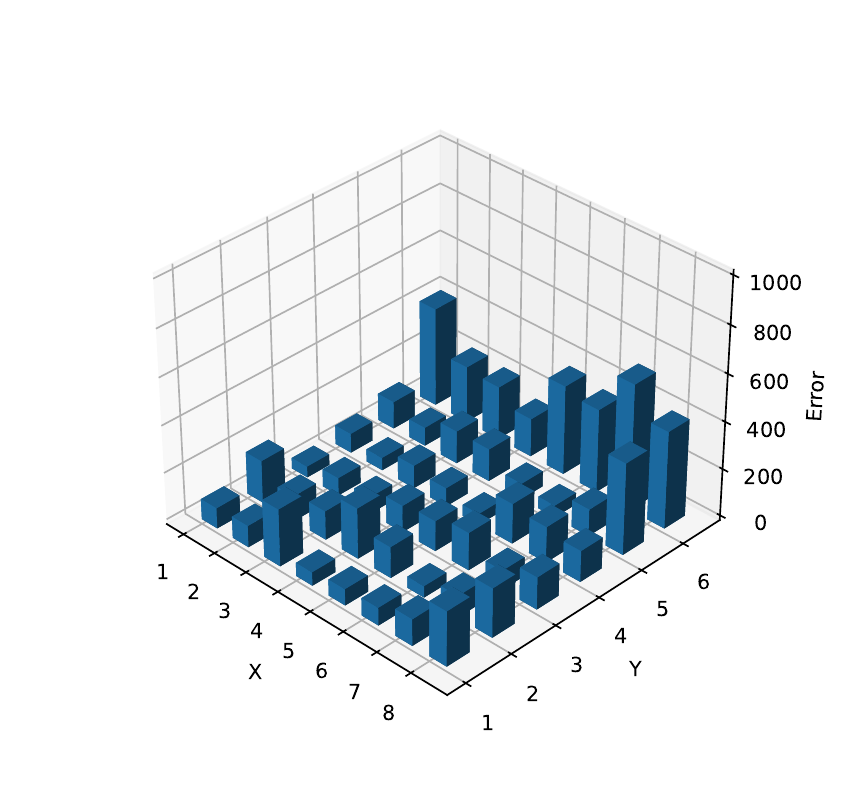}}
    \caption{3D plot obtained using \textit{EPSNoDe\textsubscript{MA}} architecture in case B with the obstacle placed within the grid map in the top-right corner. The graph shows the total error computed applying Eq. \ref{eq:toterr} for each map grid point. The highest errors are detected on the top-right corner of the map where the bars are higher.}
    \label{fig:MAall}
\end{figure}

A detailed visualization is carried out for the \textit{EPSNoDe\textsubscript{MA}} applied to case B to better understand how the framework behaves for the warning zone identification. Fig. \ref{fig:MAall} shows the total error resulting from applying Eq. \ref{eq:toterr}. 

The plot shows the framework's difficulty in reconstructing the nominal input values when close to the perturbation due to the untrained data (e.g. high error). In case B, the novelty is located in the top-right corner of the map. This obstacle introduces an alternation of LoS and NLoS conditions during the UWB measurements. As evident, the discrepancy appears near the perturbation and along the top line of the grid map. This behavior is closely tied to the reflections induced by the metal plates introduced in the environment, altering the ToF of the UWB signal. When considering the first 6 peaks of the CIR (\textit{EPSNoDe\textsubscript{MA}}), the reflections become more prominent, facilitating the framework's detection of the novelty.

\begin{figure}[t]
    \centerline{\includegraphics[width=0.9\columnwidth]{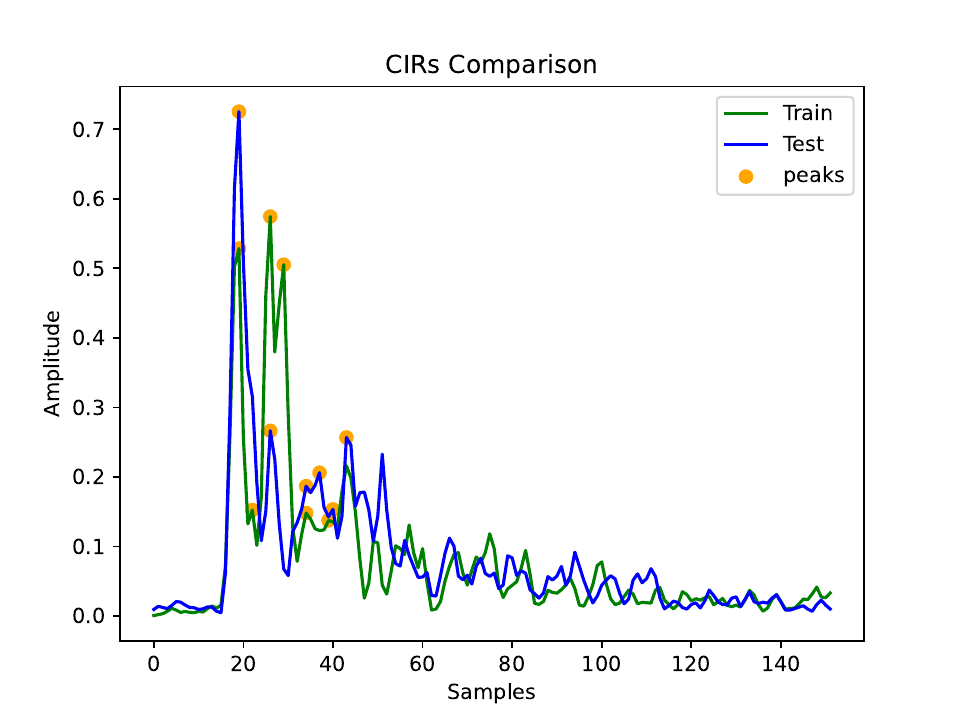}}
    \caption{Comparison between the normal CIR and the anomalous one, with both curves obtained at the same grid point. The observed discrepancies are attributed to reflections originating from a new object near that grid point.}
    \label{fig:CIRs}
\end{figure}

A clearer explanation of the previously mentioned point is presented in Fig. \ref{fig:CIRs}, which displays both the normal and anomalous CIR. Notably, the two signals look completely different from each other due to reflections caused by the new obstacle. Trained exclusively with nominal CIR signals, the model fails to reconstruct the anomalous one, leading to an output range that diverges from the real one, contributing to the global error depicted in Fig. \ref{fig:MAall}.

\begin{figure}[t]
    \centerline{\includegraphics[width=0.8\columnwidth]{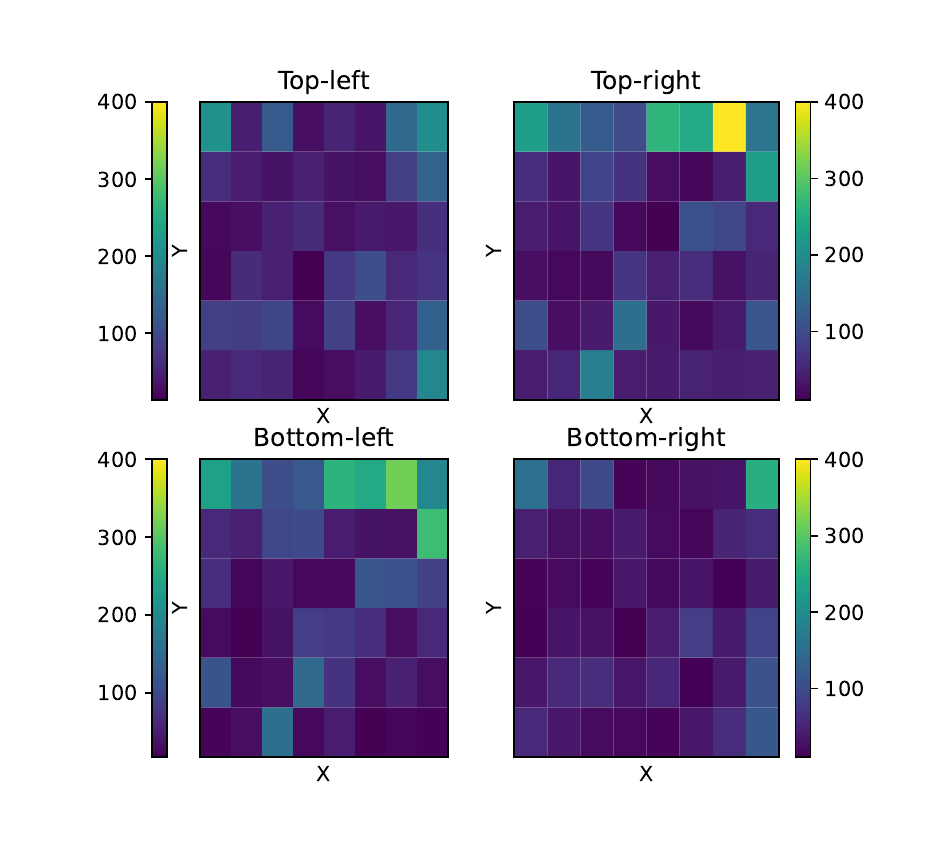}}
    \caption{The heat maps are obtained by applying Equation \ref{eq:err} for each anchor. The novelty is placed on the top-right corner of the map. The highest errors are detected from the top-right and bottom-left anchors of the map.}
    \label{fig:MA_anch}
\end{figure}

Another intriguing aspect of this generative framework is the ability to identify the anchor which mainly contributes to the total error. The representation of the error for each anchor is shown in Fig. \ref{fig:MA_anch}. In this instance, the error for each anchor is computed using Eq. \ref{eq:err}, which is the relative difference between the real and the predicted range. Notably, the top-right and bottom-left anchors significantly contribute to the total error, as shown in the heat map, suggesting the presence of an obstacle between them. By examining the anomaly grid zone, it is possible to assume that the obstacle is located in the top-right corner.

Finally, the results shown in Fig. \ref{fig:errors} are obtained by trying several hyperparameter combinations to find the best optimization for the related network. This optimization involves varying the values of neurons, batch size, and learning rate to achieve optimal performance for each model. The combinations considered are summarized in Table \ref{tab:gs}. 

\begin{table}[t]
    \centering
    \caption{Hyperparameters combinations examined for model optimization.}
    \begin{tabular}{cccccc}
        \toprule
        Architecture & E1,D1 Neuron & E2 Neuron & Batch Size & Learning Rate\\
        \midrule
        EPSNoDe\textsubscript{RNG} & \begin{tabular}{@{}c@{}}5,15, \\ 20\end{tabular} & \begin{tabular}{@{}c@{}}20,30, \\ 40\end{tabular}  & 16,32,64 & 0.001,0.01\\
        
        EPSNoDe\textsubscript{MA} & \begin{tabular}{@{}c@{}}50,55,60 \\ 65,70\end{tabular} & \begin{tabular}{@{}c@{}}70,75,80 \\ 85,90\end{tabular}  & 16,32,64 & 0.001,0.01\\

        EPSNoDe\textsubscript{PCA} & \begin{tabular}{@{}c@{}}120,125,130\\ 135,140\end{tabular} & \begin{tabular}{@{}c@{}}145,150,155 \\ 160,165\end{tabular}  & 16,32,64 & 0.001,0.01\\
        \bottomrule
    \end{tabular}
    \label{tab:gs}
    \vspace{-2mm}
\end{table}

The conducted tests involved also varying the number of training samples to explore potential limitations arising from insufficient data. Cases A and B utilize five datasets to train the model, whereas Case C employs only two. As depicted in Fig. \ref{fig:errors}, the framework demonstrates the ability to detect novelties in the environment even with only two datasets being used for training. On the other hand, increasing the number of datasets used for the training also enhances the reliability of novelty detection due to a higher overfit of the normal environment. The best hyperparameter combinations for each model are shown in Table \ref{tab:gs_best}.

\begin{table}[t]
    \centering
    \caption{Best models found after the hyperparameters evaluation.}
    
    \begin{tabular}{cccccc}
        \toprule
        Architecture & E1,D1 Neuron & E2 Neuron & Batch Size & Learning Rate\\
        \midrule
        EPSNoDe\textsubscript{RNG} & 15 & 30  & 32 & 0.001\\
        
        EPSNoDe\textsubscript{MA} & 70 & 90 & 64 & 0.001\\

        EPSNoDe\textsubscript{PCA} & 120 & 165  & 32 & 0.001\\
        \bottomrule
    \end{tabular}
    \label{tab:gs_best}
    \vspace{-2mm}
\end{table}

\subsection{Error Density Quantitative Evaluation}

In the last step, a similarity metric is employed to enhance the visualization of the fit of each model to the ground truth. To this end, the heat maps are transformed into Probability Density Functions (PDF) using the Kernel Density Estimation (KDE) method. Subsequently, the Kullback-Leibler (KL) Divergence is employed to quantify the similarity between the predicted PDF and the ground truth. The KL divergence has the following shape:

\begin{equation} \label{eq:kl}
   D_{KL} (P||Q) = \sum_{x\in X} p(x) \cdot log(\frac{p(x)}{q(x)})
\end{equation} 

where $p$ and $q$ are the two PDFs to compare. KL divergence is always non-negative, hence it can assume $D_{KL} (P||Q) \geq 0$. The results obtained by applying KL divergence metric are reported in Table \ref{tab:metrics}.

\begin{table}[t]
    \centering
    \caption{KL divergences computed for each model, comparing the predicted heat map's PDF with the ground truth's PDF.}
    
    \begin{tabular}{cccccc}
        \toprule
        Architecture & Case A & Case B & Case C \\
        \midrule
        EPSNoDe\textsubscript{RNG} & 0.15 &  0.19 & 0.14\\
        
        EPSNoDe\textsubscript{MA} & 0.50 &  0.46 & 0.31\\

        EPSNoDe\textsubscript{PCA} & 0.83 &  1.36 & 2.15\\
        \bottomrule
    \end{tabular}
    \label{tab:metrics}
    \vspace{-2mm}
\end{table}

Notably, the model that fits better is the \textit{EPSNoDe\textsubscript{RNG}}, which uses only ranging distances to detect novelties. On the other hand, the \textit{EPSNoDe\textsubscript{MA}} shows interesting results in both the metric table and heat map. While it may not offer the best match with the ground truth, it shows well where the novelty is detected in the environment, providing insights into specific areas of change. Hence, \textit{EPSNoDe\textsubscript{RNG}} can be effectively used to mitigate localization errors, while the \textit{EPSNoDe\textsubscript{MA}} to detect subtle environment changes. \textit{EPSNoDe\textsubscript{PCA}} shows the worst behavior compared to the others, indicating the inadequacy of the network architecture for this input type.

\section{Conclusions}\label{sec:conclusions}
We believe that our work will contribute to the ongoing efforts to realize the full potential of UWB technology. The objective of this study consists of advancing our understanding of the application of UWB signals in autonomous robot localization. Instead of correcting the robot's localization error through mitigation or compensation techniques, this paper focuses on mapping a specific environment to detect changes in UWB reliability within it.
Future works will see the integration of the proposed method in robot localization and navigation tasks. The UWB error information provided by our solution can serve as sensor fusion prior knowledge or as a navigation cost term.
Further investigations will be directed toward developing different approaches for detecting novelty; one example is the use of Normalizing Flow models, which distinguish themselves from Autoencoders by employing density distribution estimation for input generation, thereby facilitating the computation of the reconstruction error.

\subsection*{Acknowledgements} This work has been developed with the contribution of Politecnico di Torino Interdepartmental Center for Service Robotics PIC4SeR \footnote{\url{www.pic4ser.polito.it}}.

\bibliographystyle{unsrt}  
\bibliography{bibliography}  


\end{document}